\documentclass[10pt,twocolumn,letterpaper]{article}

\usepackage{cvpr}              
\usepackage{mathtools}
\usepackage[normalem]{ulem}
\usepackage{booktabs, multicol, multirow}

\definecolor{cvprblue}{rgb}{0.21,0.49,0.74}
\usepackage[pagebackref,breaklinks,colorlinks,allcolors=cvprblue]{hyperref}
\usepackage{multirow}

\newcommand{\ourwork}{\textbf{HierEdit }}
\usepackage{xcolor}
\newcommand{\yy}[1]{\textcolor{black}{#1}}

\def\paperID{39528} 
\def\confName{CVPR}
\def\confYear{2026}

\title{\ourwork: Region-Aware Hierarchical Diffusion for Efficient High-Resolution Editing}

\author{Yuyao Zhang\\
Dartmouth College\\
\and
Alexander Huang-Menders\\
Dartmouth College\\
\and
Yu-Wing Tai\\
Dartmouth College\\
}

\begin{document}
\twocolumn[{%
\renewcommand\twocolumn[1][]{#1}%
\maketitle
\centering
\vspace{-0.15in}
\includegraphics[width=.96\textwidth]{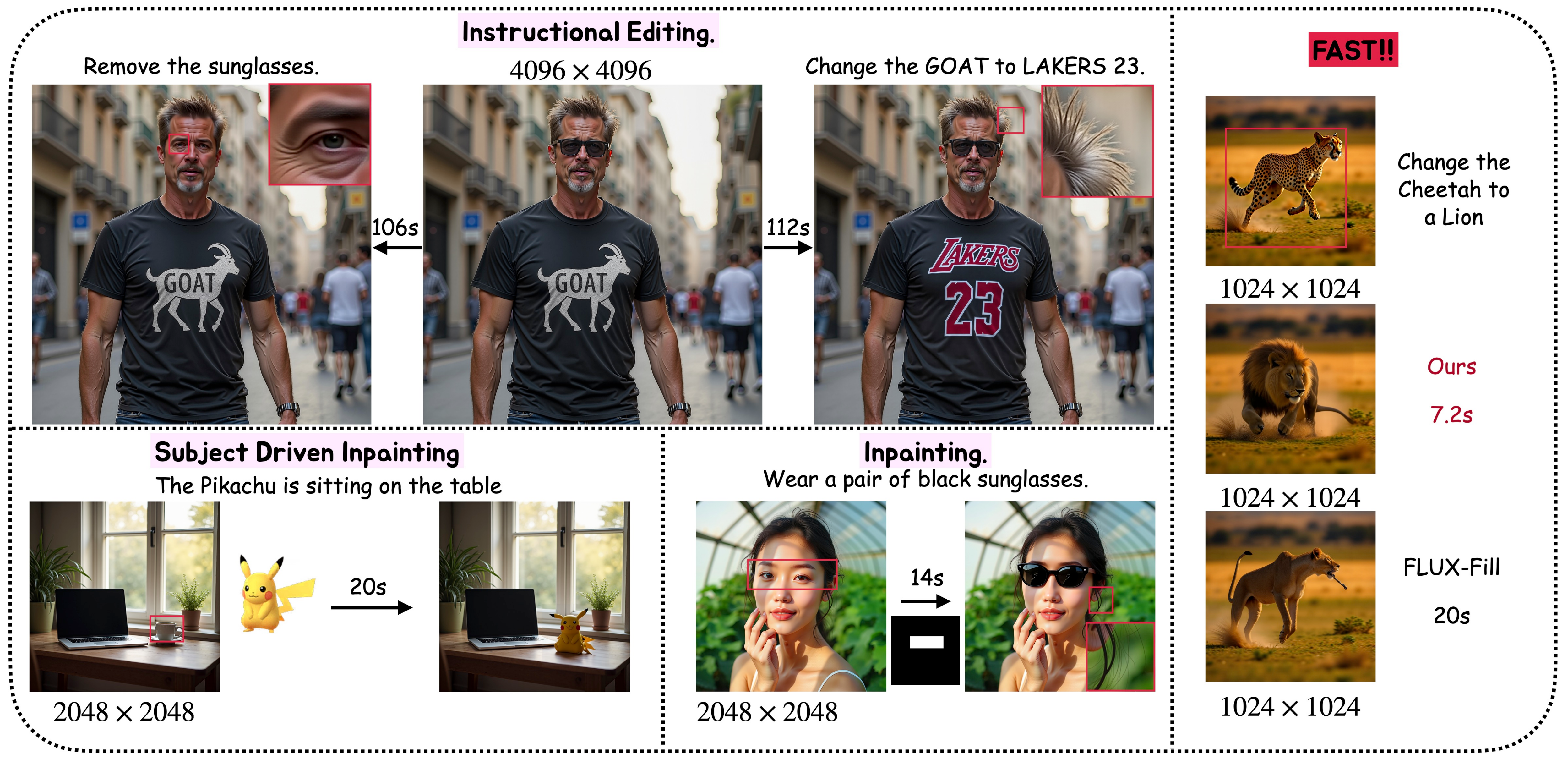}\\
\vspace{-1.0em}
\captionof{figure}{
    \textbf{Demo of \ourwork.} Our method enables efficient, high-fidelity local editing
    at ultra-high resolution (2K and 4K here) without the need of 4K data for training. We support both instruction editing and both text- and image-guided inpainting. On commodity resolution (1K) we are much faster.
}
\label{fig:teaser}
\vspace{0.5em}
}]

\begin{abstract}
\vspace{0.2mm}
High-resolution image editing is essential for professional and creative applications, yet existing multimodal diffusion-based editors remain computationally inefficient and constrained to relatively low resolutions. Current approaches redundantly process the entire image canvas or rely on large-scale high-resolution datasets, resulting in substantial training and inference costs. We introduce \ourwork, a region-aware hierarchical diffusion framework designed for efficient and scalable high-resolution image editing. Our method first performs edits on a low-resolution proxy using an off-the-shelf editing model to generate a reference and to localize the modified regions. A hierarchical local-window diffusion model (\textbf{Local-Window MMDiT}) that refines only edited regions within the original high-res image, while reusing the unaltered regions as conditioning inputs. The low-resolution proxy further provides structural guidance and intermediate denoising supervision (\textbf{Inference Acceleration}) , ensuring consistent global semantics and stable generation without the need for full-resolution attention computation. This targeted and hierarchical design enables fast, high-fidelity editing of images up to 4K resolution without any specialized high-resolution training data. Extensive experiments demonstrate that \ourwork achieves competitive visual quality on commodity-resolution datasets while significantly accelerating inference and extending seamlessly to ultra-high-resolution 4K editing. Please check our {\href{https://peteryyzhang.github.io/HierEdit-page/}{\textbf{Project Page}}}.
\end{abstract}
    
\section{Introduction} 
\label{sec:intro} 

Text-guided image editing has become a powerful tool for creative expression and content manipulation, enabling users to modify images with precise semantic control~\cite{rombach2022high, podell2023sdxl, blackforestlabs2024flux1dev}. However, the computational cost of dense attention~\cite{dao2022flashattention, dao2023flashattention} scales quadratically with resolution, limiting their use in high-resolution workflows. Most state-of-the-art techniques remain constrained to standard resolutions, typically below $1K \times 1K$, which restricts their applicability in professional domains such as digital advertising, cinematography, and high-fidelity visualization, where ultra-high-resolution (UHR) outputs (4K and above) are essential.


Image editing tasks can be broadly categorized into \textit{global} edits, like applying a new artistic style, which require processing the entire image, and \textit{local} edits, which target specific objects or regions, such as ``Remove the rightmost person.'' or ``Replace an apple with an orange.'' Many practical applications fall into the latter category, yet existing models handle them inefficiently. Some re-render the entire image for minor changes, wasting computation on untouched regions, while others rely on inpainting, which ignores interactions with external areas and often produces border artifacts. The key challenge is not just scaling to higher resolutions, but enabling targeted local edits efficiently. Recent hierarchical local attention architectures show promise for efficient \textit{generation}~\cite{peebles2023scalable}, but their use in editing remains largely unexplored.

In this work, we introduce \ourwork, a novel framework for efficient, high-resolution local image editing. To avoid costly dense attention across the entire image, we employ \textbf{Local-Window MMDiT}, which performs hierarchical local-window attention and surgically targets only regions requiring modification. The pipeline begins by downsampling the high-resolution input to a standard-resolution proxy (e.g., $1$K $\rightarrow 256$), where an off-the-shelf editing model~\cite{flux2024} generates edits. This low-resolution proxy serves two roles: (1) providing {semantic guidance and a refined edit mask} to localize modifications accurately, and (2) serving as an {intermediate initialization} to skip early denoising steps. Guided by the proxy, the Local-Window MMDiT then re-synthesizes only the masked patches in the original high-resolution image, preserving global consistency and pixel fidelity. Finally, \textbf{inference acceleration} techniques, including intermediate flow initialization and kernel optimizations, further reduce runtime while enabling ultra-high-resolution editing.

This targeted methodology achieves substantial gains in computational efficiency, memory usage, and inference speed, particularly for common local editing tasks. By restricting expensive dense attention operations to sparse local-window attention on a small subset of the image, \ourwork\ decouples performance from image resolution, enabling rapid, iterative edits that would be impractical with dense-attention models. A direct benefit of this efficiency is high scalability, allowing precise editing of ultra-high-resolution (UHR) images such as 4K. Our framework thus provides a practical and generalizable solution for high-fidelity local editing. 

\vspace{2mm}
\noindent\textbf{Our main contributions are summarized as follows:}

\begin{itemize} 
    \item We introduce a low-resolution edited proxy to guide high-resolution local editing, providing semantic guidance, a refined edit mask, and an intermediate initialization for skipping early denoising steps, achieving up to \yy{$>6\times$ faster performance in $1$K resolution (more gains at higher resolutions)} with state-of-the-art quality. 
    \item We propose \textbf{Local-Window MMDiT}, a hierarchical diffusion model that replaces dense attention with sparse local-window attention, operating only on regions requiring modification, dramatically improving computational efficiency.
    \item We incorporate \textbf{Inference Acceleration} techniques, including intermediate flow initialization and kernel-level optimizations, further reducing runtime for high-resolution editing.
    \item Our framework enables fast, high-fidelity editing on ultra-high-resolution (4K) images without requiring 4K training data, making practical UHR editing feasible for professional workflows.
\end{itemize}

\begin{figure*}[htbp]
    \centering
    \includegraphics[width=.95\linewidth]{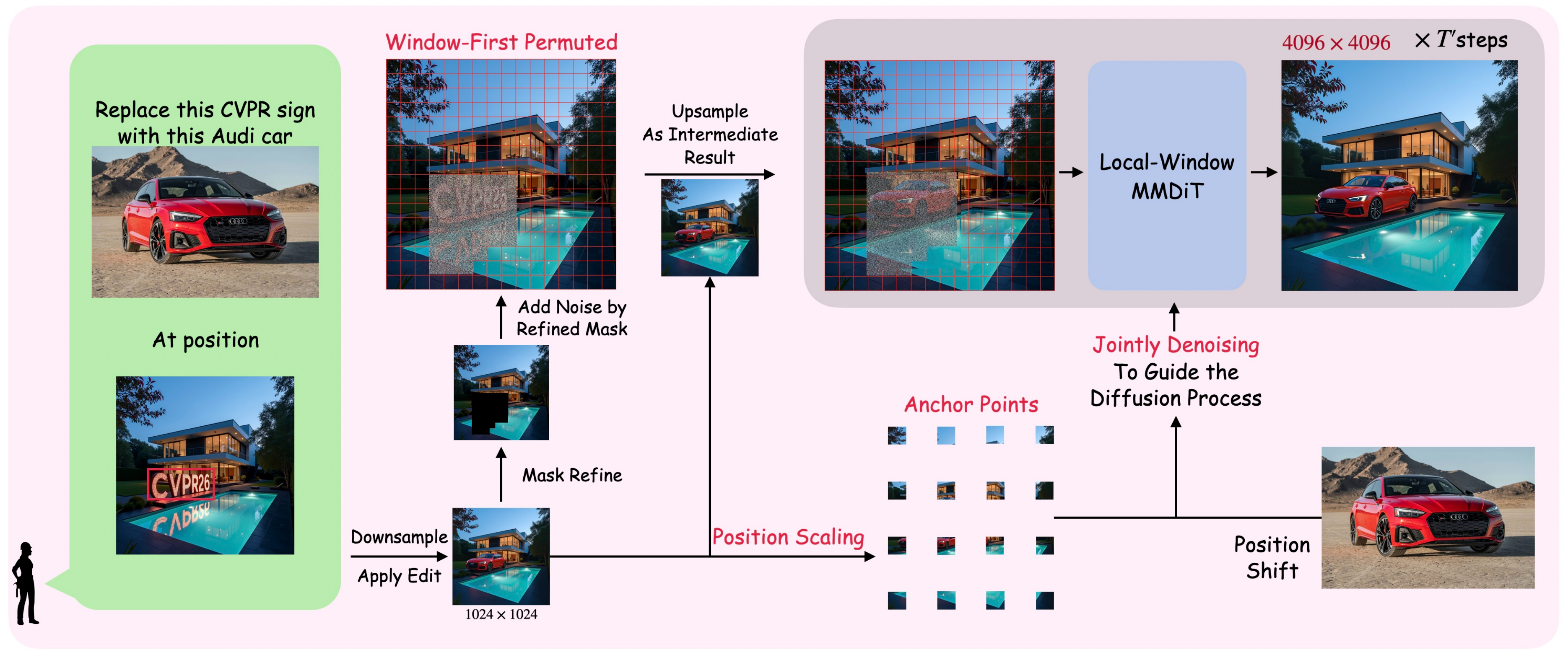}\\
    \vspace{-0.1in}
    \caption{Schematic of the \ourwork \space framework. We employ editing of a downsampled input image and region bounding to identify the edited patches and obtain the low-resolution proxy. We then proceed to input concatenation, re-permutation, and positional encoding. Finally, we pass this input to our hierarchical local-window MMDiT model, which generates the high resolution edited results with less denoising steps as upsampled low-resolution proxy can serve as intermediate denosing result.}
    \vspace{-0.2in}
    \label{fig:diagram}
\end{figure*}
\section{Related Work}
\noindent\textbf{Controllable Image Generation.}
\yy{Text-to-image (T2I) generation has advanced rapidly with diffusion models, progressing from pixel-space methods such as GLIDE~\cite{nichol2022glide} and Imagen~\cite{saharia2022photorealistic} to efficient latent-space frameworks such as Stable Diffusion~\cite{rombach2022high} and Raphael~\cite{xue2023raphael}. Improvements in multimodal alignment, e.g., DALLE-2~\cite{ramesh2022hierarchical} and Playground~\cite{li2024playground}, and in architecture design, e.g., Diffusion Transformers~\cite{peebles2023scalable}, PixArt~\cite{chen2024pixart}, and FLUX~\cite{flux2024}, have further improved image fidelity and diversity. Recent systems~\cite{ding2021cogview,rombach2022high,chen2023pixart,podell2023sdxl,peebles2023scalable,xue2023raphael,li2024hunyuan,chen2024pixart,li2024playground,blackforestlabs2024flux1dev,cai2025hidream,xie2024sana,xie2025sana,gao2025seedream} adopt cross-attention and diffusion transformer (DiT) architectures to achieve near-photorealistic synthesis at resolutions up to $1\text{K}\times1\text{K}$, forming a strong foundation for controllable generation.}

\yy{Despite this progress, fine-grained spatial and semantic control remains challenging. Personalization methods~\cite{ruiz2023dreambooth, gal2022image} enable subject-specific generation but typically require instance-level fine-tuning. Structured control methods such as ControlNet~\cite{zhang2023adding} and GLIGEN~\cite{li2023gligen} provide spatial conditioning through edge maps or bounding boxes, but rely on detailed external guidance and offer limited holistic scene reasoning. Domain-specialized methods such as Raphael~\cite{xue2023raphael} improve fidelity at high computational cost, while lightweight adapters such as Attend~\cite{chefer2023attend} reduce overhead but remain limited on complex compositions. Autoregressive frameworks, including LlamaGen~\cite{sun2024autoregressive}, Show-O~\cite{xie2024showo}, and Janus-Pro~\cite{chen2025janus}, support prompt-driven synthesis but often suffer from weaker spatial consistency. Overall, most existing approaches are still constrained by quadratic attention and limited high-resolution training data, hindering scalable ultra-high-resolution synthesis and region-aware editing.
}

\vspace{2mm}
\noindent\textbf{High-Resolution Image Synthesis.} Real-world applications increasingly demand $4K$ or higher resolutions, driving research beyond the $1K \times 1K$ barrier. Works such as PixArt-$\Sigma$ and SANA 1.5~\cite{xie2025sana} achieve near-$4K$ synthesis via extensive high-resolution pretraining, while others~\cite{hoogeboom2023simple,liu2024linfusion,ren2024ultrapixel,xie2023difffit,teng2023relay,zheng2024any,zhang2025diffusion,yu2025ultra} rely on fine-tuning or training-from-scratch with curated datasets. For instance,~\cite{zhang2025diffusion} employs wavelet supervision to enhance detail clarity, and~\cite{yu2025ultra} introduces lightweight fine-tuning for higher-resolution adaptation. Despite their effectiveness, these methods remain limited by the scarcity of high-quality high-resolution data and high GPU requirements. \yy{One work~\cite{zhang2025scale} demonstrates a promising direction by leveraging scaled RoPE embeddings and window-permuted global-local attention to scale pretrained diffusion models beyond their pretrained resolutions; we build on this idea in the architecture design of our method.}

More recent training-free strategies~\cite{guo2024make,qiu2024freescale,du2024max,liu2024hiprompt,wu2025megafusion,shi2025resmaster,kim2025diffusehigh,huang2024fouriscale,bu2025hiflow} avoid data collection by leveraging pretrained models directly. 
While promising, these approaches often inherit significant runtime and memory overhead, limiting accessibility for broader use.

\vspace{2mm}
\noindent\textbf{Attention Acceleration.}
As image and video resolutions grow, quadratic attention becomes the main bottleneck. Linear-attention variants like the SANA series~\cite{xie2024sana,xie2025sana,zhu2025dig} reduce complexity but suffer from non-injective mappings and lower attention spikiness~\cite{han2024bridging,meng2025polaformer,zhang2402hedgehog}, often causing local detail inconsistencies. System-level optimizations such as FlashAttention~\cite{dao2022flashattention,dao2023flashattention,shah2024flashattention}, quantization~\cite{zhang2025sageattention,zhang2024sageattention2}, and sparsity-based methods~\cite{deng2024attention,liu2022dynamic} improve efficiency via hardware-aware acceleration. Architectures like LongFormer~\cite{beltagy2020longformer} and SwinFormer~\cite{liu2021swin} use local windows but still process the full image with static attention. Recent works~\cite{lai2025flexprefill,zhang2025spargeattn,xu2025xattention,xi2025sparse,yang2025sparse,yuan2024ditfastattn,zhang2025ditfastattnv2} employ block-sparse or compressed attention in diffusion transformers, achieving moderate speedups ($1.5$–$1.8\times$) at some cost to fine-grained fidelity. Complementary methods like OminiControl2~\cite{tan2025ominicontrolminimaluniversalcontrol,tan2025ominicontrol2efficientconditioningdiffusion} speed up conditional diffusion by restricting computation to masked regions but still rely on user-provided masks and predefined edit scopes. In contrast, our approach automatically localizes editable regions using low-resolution proxies and applies content-aware sparse attention, enabling efficient ultra-high-resolution editing without manual masks or retraining.
\section{Method}
\label{sec:method}
Figure~\ref{fig:diagram} presents an overview of the proposed \textbf{\ourwork} framework, which leverages low-resolution edits to guide high-resolution image editing efficiently. Given a high-resolution input image, a text prompt, and optional control inputs, we first downsample the image and apply edits using a state-of-the-art (SOTA) image editing model. Comparing the edited low-resolution output with the original image produces a refined mask that accurately localizes the edited regions. These refined regions, together with the low-resolution edits, serve as references for high-resolution, region-aware generation.
 
In Section~\ref{sec:pre}, we review the foundational preliminaries, including rectified flow models and the Multi-Modal Diffusion Transformer (MMDiT) backbone with RoPE position encoding. Section~\ref{sec:lowres} describes low-resolution guided region refinement, where the refined mask and edited low-resolution image are computed to guide subsequent high-resolution editing. Section~\ref{sec:mmdit} introduces Local-Window MMDiT, which combines efficient local-window attention, joint denoising of integrated token sequences, and low-resolution anchors to maintain global semantic coherence while reducing computation. Finally, Section~\ref{sec:acc} presents inference acceleration technique with intermediate flow initialization to further speed up high-resolution generation.

\subsection{Preliminaries}\label{sec:pre}
\noindent\textbf{Rectified Flow Models~\cite{liu2022flow}.} Flow-based generative models learn a continuous transformation between a data sample $X_0$ and a simple reference distribution $X_1$ by integrating an ODE
\[
\frac{dZ_t}{dt} = V(Z_t, t),
\]
where $V$ is a learned velocity field. Sampling is performed by drawing $Z_1 \sim \mathcal{N}(0,I)$ and solving the ODE backwards to obtain $Z_0$.
Rectified flows constrain the intermediate states to follow a linear path,
\[
Z_t = (1 - t)X_0 + tX_1,
\]
which produces straight and stable trajectories that require only a small number of integration steps. 

\vspace{2mm}
\noindent\textbf{MMDiT with RoPE Position Encoding.} The Multi-Modal Diffusion Transformer (MMDiT) is the standard backbone of recent text-to-image models~\cite{flux2024, cai2025hidream, gao2025seedream}. It processes a text token sequence $C_T$ and a noisy image token sequence $X$, which are concatenated as $[C_T; X]$ and jointly encoded through a unified Multi-Modal Attention (MMA) layer. 
Spatial structure in image tokens is represented using Rotary Position Embeddings (RoPE)~\cite{su2024roformer}, applying a position-dependent rotation $\mathrm{Rot}(i,j)$ to each spatial token:
\[
X_{i,j} = X_{i,j} \cdot \mathrm{Rot}(i,j).
\]
The MMA mechanism performs standard self-attention over the combined sequence:
\[
\mathrm{MMA}([C_T; X]) =
\mathrm{Softmax}\!\left(
\frac{QK^{\top}}{\sqrt{d}}
\right)V,
\]
where $Q, K, V$ denote the projection matrices for both text and image tokens. This formulation enables dense bidirectional interaction across modalities. The usage of RoPE enables the model to generate image of sizes within a pretrained range, regardless of where it is on the latent grid.
\begin{figure}[h]
    \centering
    \includegraphics[width=0.95\linewidth]{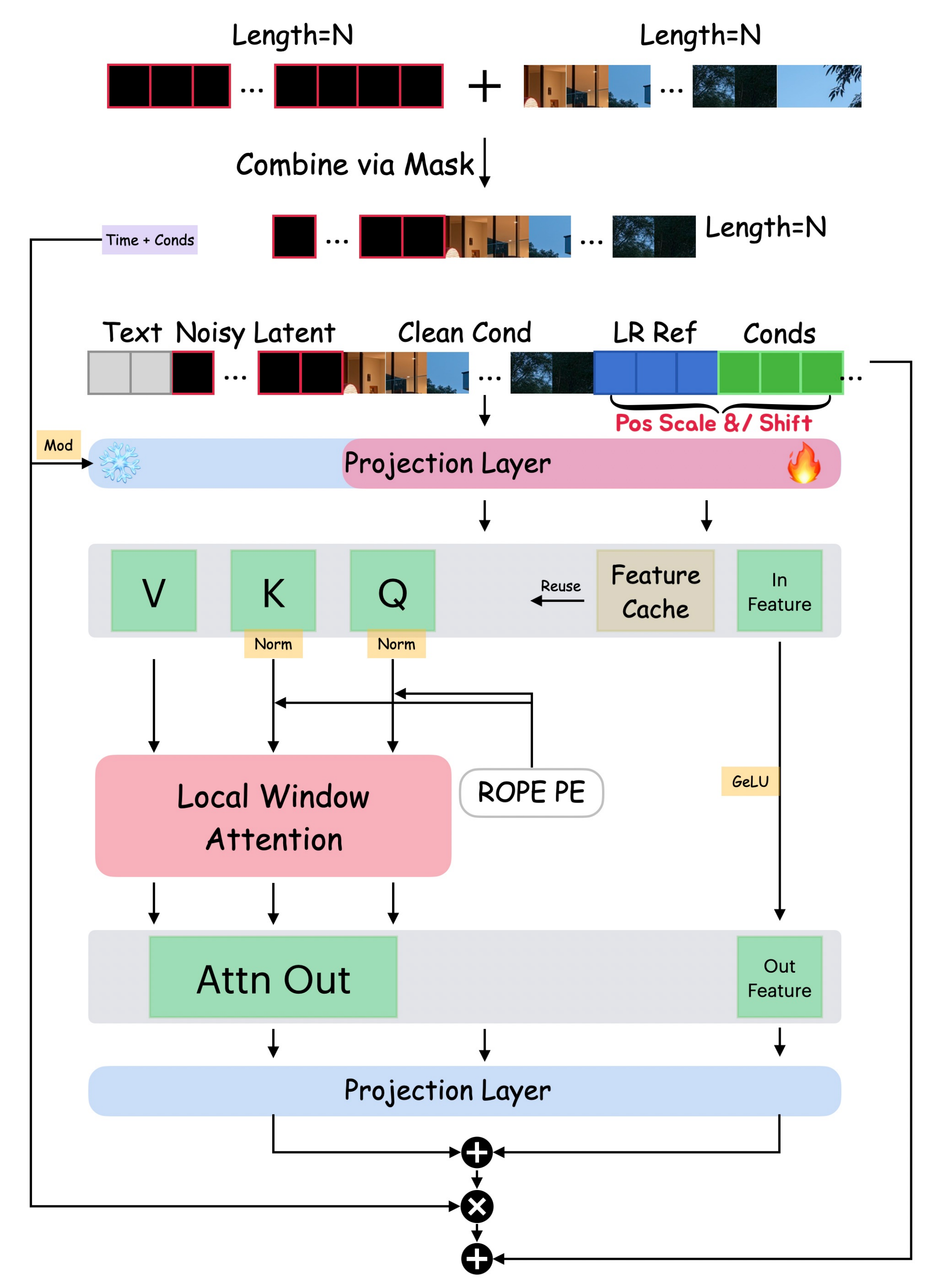}\\
    \vspace{-0.1in}
    \caption{The illustration of jointly denoising the integrated token sequence via Local-Window MMDiT. We prune the tokens from 2N to N by the edit mask and only denoise the edited region. All tokens are concatenated together while conditions and low-resolution reference are processed via finetuned layers, and original weights for the image latent. "Fire" and "Snowflake" signs means LoRA finetuning and frozen weights.}
    \vspace{-0.15in}
    \label{fig:joint-denoise}
\end{figure}

\subsection{Low-Resolution Guided Region Refinement}
\label{sec:lowres}

\yy{Given a high-resolution input image $X_{Hr}$, a text prompt, and an optional control image $X_{Control}$, we first downsample $X_{Hr}$ to $X_{Lr}$ (e.g., $1\text{K}\times1K\text{K}\rightarrow256\times256$). We then feed $X_{Lr}$, together with the prompt and optional control input, into a unified image generation and editing model~\cite{flux2024} to perform text-guided or multimodal conditional editing, producing an edited low-resolution image $X_{Lr}’$ that guides the subsequent high-resolution editing branch.}

\yy{Since user-provided masks $M$ are often imperfect for complex effects such as reflections, shadows, and occlusions, we derive a refined mask $\tilde{M}$ by comparing $X_{Lr}’$ with $X_{Lr}$ at the pixel level to better localize the edited regions. Overall, the pipeline downsamples $X_{Hr}$ to $X_{Lr}$, obtains $X_{Lr}’$ through low-resolution editing, and computes $\tilde{M}$ for precise high-resolution editing. This refined mask ensures better alignment between the final high-resolution edits and the intended semantic and contextual modifications.}

\subsection{Local-Window MMDiT}\label{sec:mmdit}

\noindent\textbf{Efficient Local Window Attention.}
DiT-based generative models rely heavily on self-attention, whose quadratic computational cost becomes prohibitive as image size increases. For an image of size $H \times W$ after encoder downsampling, the number of spatial tokens is $N = H \cdot W$, resulting in $O(N^2)$ complexity. To address this, we partition the high-resolution latent $X \in \mathbb{R}^{H \times W}$ into non-overlapping windows $x_i$ of size $l$, bounded by the pretrained resolution (e.g., 1024), with $i = \frac{H}{l} \times \frac{W}{l}$ windows in total. Attention is computed only within each window, reducing complexity to $O\big(\frac{H}{l} \cdot \frac{W}{l} \cdot (l^2)^2\big) = O(N \cdot l^2)$. For example, scaling from $1024\times1024$ to $4096\times4096$ reduces computation by $256\times$.

In practice, we set $l = 16$, corresponding to 256 pixels in the original image, balancing accuracy and efficiency: smaller windows speed up attention but overly small windows underutilize GPU kernels. Importantly, while this analysis assumes processing all windows, in practice we compute attention only for masked local windows requiring edits. As a result, runtime scales linearly with the number of local windows being processed, further accelerating high-resolution generation. Because each local window has a fixed positional embedding range, the model can reliably generate content within each block, enabling ultra-high-resolution synthesis. \yy{To prevent block artifacts introduced by the non-overlapping windows, we allow each window to attend to boundary tokens of adjacent windows, enabling information flow among them without breaking the linearized computational burden.}

\vspace{2mm}
\noindent\textbf{Jointly Denoising Integrated Token Sequence.}
In text- or image-guided editing tasks, the model typically processes two input sequences of equal length: the masked source image $X_{\text{origin}}$ and its corresponding noisy latent $X_{\text{noise}}$. Naively concatenating these doubles the sequence length and quadruples attention memory cost.


\yy{To address this, we construct a unified token sequence by assigning each spatial location a single role. Unmasked regions from $X_{\text{origin}}$, denoted $C^{\text{mask}=0}$, serve as \emph{condition tokens} that provide static contextual information and are not denoised, while masked regions from $X_{\text{noise}}^{\text{mask}=1}$ are \emph{noisy tokens} that undergo iterative denoising. Combining these into a single sequence $X_{\text{integrated}} = [\,C^{\text{mask}=0};\; X_{\text{noise}}^{\text{mask}=1}\,]$ reduces the total length to that of a single image, preserving computational efficiency. Because condition tokens remain fixed across diffusion steps, their Key and Value projections can be cached and reused \yy{after one single forward pass(Feature Caching)}, significantly accelerating inference. To prevent noise contamination, condition tokens attend only to other condition tokens and not to the noisy tokens or text tokens.}
\yy{To better preserve the details in the condition images, our framework incorporates additional guidance signals by concatenating them as extra tokens along the sequence dimension. All conditions interact with the generation tokens through the self-attention layers of the pretrained DiT, allowing us to leverage the original model weights without architectural modification; only lightweight LoRA modules are fine-tuned to adapt the model to the new conditioning scheme. }
Figure~\ref{fig:joint-denoise} illustrates the overall architecture. Speed improvements are demonstrated in the ablation study.
\vspace{2mm}

\noindent\textbf{Maintaining Global Semantics via Low-Res Anchors.}
Partitioning a high-resolution image into independent local windows can introduce discontinuities and disrupt global semantic coherence. We observe that long-range dependencies primarily govern global structure and spatial layout, while fine-grained details depend on local interactions. To preserve global consistency, we leverage a pre-edited low-resolution image $X_{Lr}' \in \mathbb{R}^{h \times w}$, which encapsulates holistic contextual information. Pixel positions in this image are indexed as $(m, n)$ with $m \in {0, \dots, h-1}$ and $n \in {0, \dots, w-1}$.

We define a scaling ratio $\rho = \frac{H}{h}$ (empirically set to 4) and map low-resolution coordinates to high-resolution space as $(\tilde{m}, \tilde{n}) = (\rho \cdot m, \rho \cdot n)$. Each projected token $X_{Lr}'[\tilde{m}, \tilde{n}]$ provides contextual cues for the corresponding high-resolution region. Reference image tokens are similarly scaled and shifted, $(m', n') = \rho' \cdot ((m', n') + \Delta)$, where $\rho'$ is the size ratio and $\Delta$ the position offset.

We then construct a unified token sequence $[C_T, X_{\text{integrated}}, X_{Lr}', X_{\text{control}}, \dots]$ and process all conditions with the standard DiT encoders the base model use. After scaling and shifting position embeddings, LoRA only adapts the attention projection layers to the new features. To prevent noisy latents from affecting conditions, only self-attention is applied to the condition tokens.

Training uses a standard flow-matching loss, which aims at teaching model about the new attention pattern. This design provides three key benefits: (1) training efficiency by updating only a compact set of LoRA parameters; (2) adaptation using commodity-resolution data (1K resolution); and (3) seamless extension to ultra-high-resolution generation (e.g., 4K) since local-window attention focuses on pretrained-size regions, making it resolution-invariant.

\begin{figure*}[h]
    \centering
    \includegraphics[width=.95\linewidth]{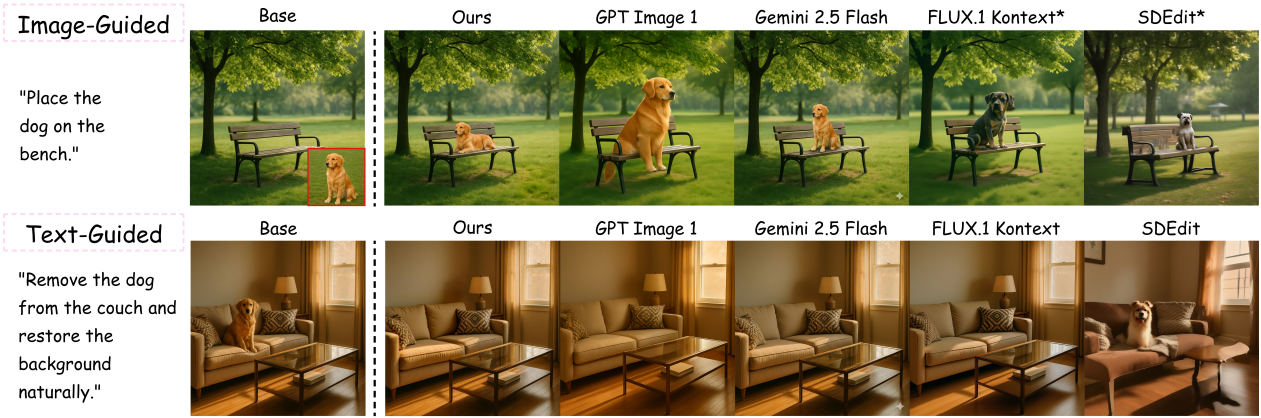}\\
    \vspace{-0.1in}
    \caption{Qualitative comparison for instructional editing. The asterisk \textbf{*} indicates that the model does not take a control image. We demonstrate more accurate subject integration by not only preserving subject identity, but also integrating it more naturally with the scene context. We also match the best lighting preservation.}
    \vspace{-0.1in}
    \label{fig:main-qualitative-instructional}
\end{figure*}

\begin{figure*}[h]
    \centering
    \includegraphics[width=.95\linewidth]{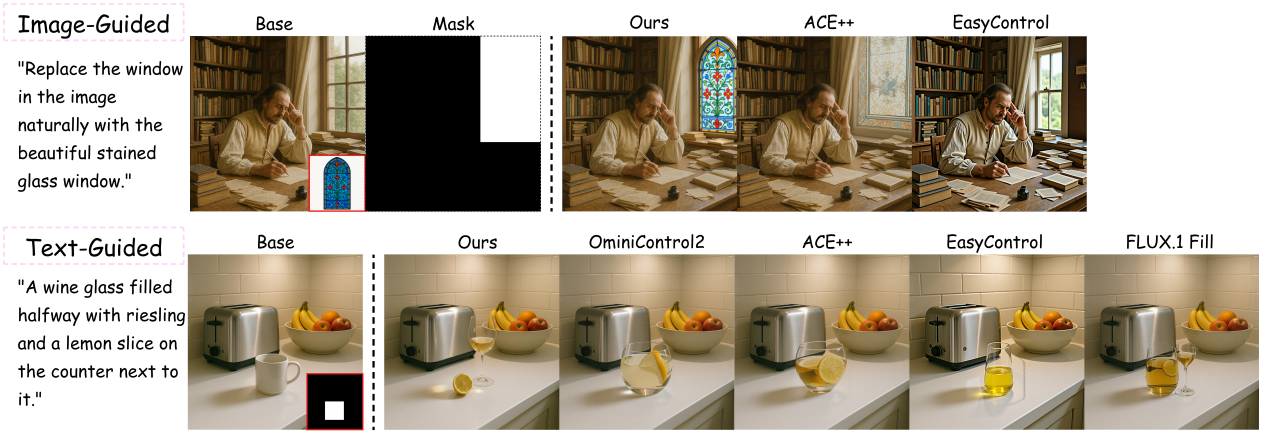}\\
    \vspace{-0.1in}
    \caption{Qualitative comparison of inpainting-based editing methods. The proposed HierEdit demonstrates better performance in preserving unmodified regions while maintaining natural integration of edited content.}
    \vspace{-0.2in}
    \label{fig:main-qualitative-inpainting}
\end{figure*}

\subsection{Inference Acceleration}\label{sec:acc}
\noindent \textbf{Intermediate Flow Initialization.} 
As we already have the low-resolution reference, which has the same structure as the high resolution results, the only difference is the high frequency details. Therefore we first resize the low-resolution reference to the same size as the target image, sharpen it, and scale noise to the intermediate timestep $t$ to obtain $X_{\text{ref}}^{t}$. 
For a given flow, the sampling of high-resolution generation starts from the noisy variant of $X_{\text{ref}}^{t}$, which can be expressed as:
\begin{equation*}
X_{\text{hr}}^t = \alpha X_{\text{hr}}^{1} + (1 - \alpha) X_{\text{ref}}^{t},
\label{eq:init-align}
\end{equation*}
where $X_{\text{hr}}^{t}$ is the initialization of the high-resolution sampling, $X_{\text{hr}}^{1}$ is Gaussian noise, and $\alpha \in (0,1)$ is the noise addition ratio. 
Such initialization allows skipping the early stage in high-resolution generation by providing the low-frequency components. Thus we facilitate faster inference by reducing redundant denoising steps from $T\text{ to } T'$, empirically $T=28$ and $T'=10$.


\section{Experiments}
\label{sec:experiments}

\begin{figure*}[h]
    \centering
    \includegraphics[width=\linewidth]{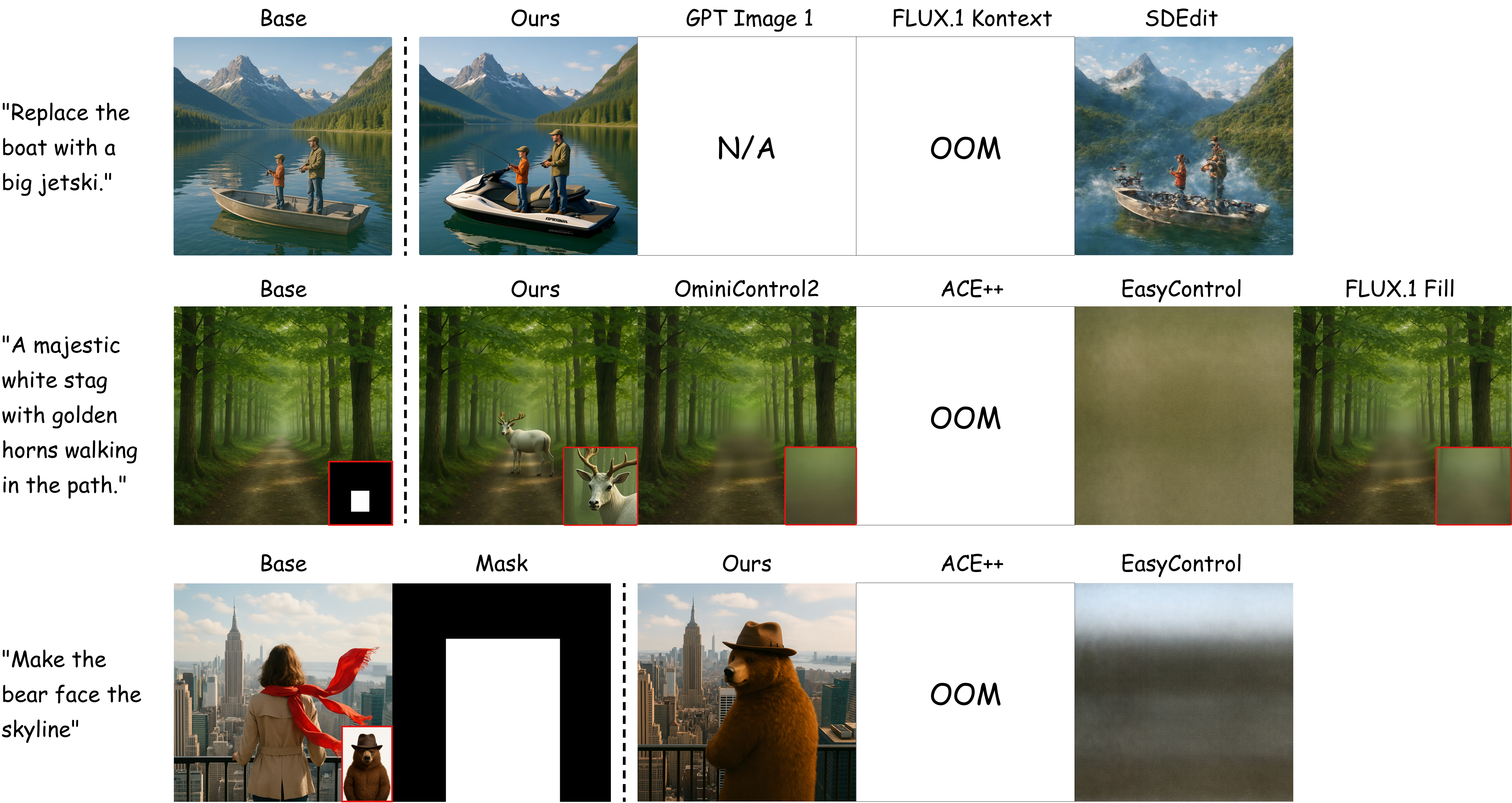}
    \vspace{-0.1in}
    \caption{Comparison at 4K resolution for text guided editing, text-guided inpainting, and subject-guided inpainting. GPT-Image-1 was incapable of generating resolutions higher than 1K by design and ACE++ was unable to run 4K inference with 96GB of GPU memory. Our method succeeded in generating the results while the rest of others failed.}
    \label{compare-res}
    \vspace{-0.15in}
\end{figure*}

\begin{table*}[h]
\centering
\small
\renewcommand{\arraystretch}{1.1}
\setlength{\tabcolsep}{6pt}
\begin{tabular}{l | l | c c | c c | c | c}
\toprule
&& \multicolumn{2}{c}{\textbf{CompBench}} & \multicolumn{2}{|c|}{\textbf{EmuEdit}} & \textbf{ImgEdit} & \textbf{I2EBench} \\
\textbf{Task} & \textbf{Method} & CLIP$\uparrow$ & SSIM$\uparrow$ & $\text{CLIP}_{\text{dir}}\uparrow$ & DINO$\uparrow$ & Composite$\uparrow$ & SSIM$\uparrow$ \\
\midrule
\multirow{4}{*}{\parbox{2.6cm}{\centering Text-guided\\editing}} 
& SDEDIT & 18.5 & 0.351 & 0.053 & 0.159 & 1.46 & 0.355 \\
& FLUX.1 Kontext & \textbf{20.8} & \textbf{0.954} & 0.116 & \textbf{0.840} & 3.45 & \underline{0.501} \\
& GPT-Image-1 & 18.9 & 0.191 & \textbf{0.132} & 0.697 & \textbf{4.45} & 0.478 \\ 
& \textbf{Ours} & \underline{20.6} & \underline{0.949} & \underline{0.117} & \underline{0.833} & \underline{3.51} & \textbf{0.508} \\
\bottomrule
\end{tabular}
\vspace{-0.1in}
\caption{Quantitative comparison of text guided image editing methods across CompBench, EmuEdit, ImgEdit, and I2EBench benchmarks.}
\vspace{-0.15in}
\label{tab:Instruction-edit_full}
\end{table*}

\begin{table*}[h]
\centering
\small
\renewcommand{\arraystretch}{1.1}
\setlength{\tabcolsep}{6pt}
\begin{tabular}{l| l| c c c  c |c c}
\toprule
\textbf{Task} & \textbf{Method} & \textbf{FID}$\downarrow$ & \textbf{PSNR}$\uparrow$ & \textbf{CLIP-T}$\uparrow$ & \textbf{CLIP-I}$\uparrow$ & \textbf{Latency}$\downarrow$ & \textbf{Time/iter}$\downarrow$ \\
\midrule
\multirow{5}{*}{\parbox{2.6cm}{\centering Text-guided inpainting}} 
 & FLUX-Fill & 56.1 & \underline{19.23} & 0.338 & 0.923 & 21.4s & 0.42s \\
 & OminiControl2* & \textbf{39.2} & 19.11 & 0.339 & 0.921 & \underline{8.25s} & \underline{0.29s} \\
 & ACE++ & 37.2 & 18.81 & \textbf{0.342} & \textbf{0.929} & 22.5s & 0.80s \\
 & EasyControl & 108.6 & 15.38 & 0.331 & 0.887 & 14.5s & 0.55s \\
 & \textbf{Ours} & \underline{39.5} & \textbf{19.31} & \underline{0.340} & \underline{0.926} & \textbf{6.97s} & \textbf{0.24s} \\
\midrule
\multirow{3}{*}{\parbox{2.6cm}{\centering Image-guided inpainting}} 
 & ACE++ & \underline{42.5} & \textbf{16.21} & \underline{0.346} & \textbf{0.959} & 52.9s & 1.9s \\
 & EasyControl & 108.3 & 14.89 & 0.342 & 0.932 & \underline{17.3s} & \underline{0.65s} \\
 & \textbf{Ours} & \textbf{41.9} & \underline{15.82}& \textbf{0.349} & \underline{0.939} & \textbf{7.84s} & \textbf{0.26s} \\
\bottomrule
\end{tabular}
\vspace{-0.1in}
\caption{Quantitative comparison of text-guided and image-guided FLUX-based inpainting methods across fidelity, perceptual, and efficiency metrics on $1K\times1K$ resolution. }
\vspace{-0.15in}
\label{tab:inpainting_comparison_full}
\end{table*}

\begin{table}[h]
\centering
\small
\renewcommand{\arraystretch}{1.1}
\setlength{\tabcolsep}{5pt}
\begin{tabular}{l l c c c c}
\toprule
\textbf{Edit \%} & \textbf{Method} & \textbf{1K} & \textbf{2K} & \textbf{3K} & \textbf{4K} \\
\midrule
\multirow{5}{*}{$25\%$}
 & OminiControl2 & \underline{5.98s}  & \underline{21.5s}  & \underline{63.3s} & \underline{155s}  \\
 & FLUX-Fill & 21.4s & 113s & 383s  & 1064s \\
 & ACE++ & 25.1s & 424s & --- & --- \\
 & EasyControl & 27.5s & 69.2s & 222s & 590s \\
 & \textbf{Ours} &\textbf{4.51s}  & \textbf{15.6s} & \textbf{35.6s}  & \textbf{91.4s} \\
\midrule
\multirow{5}{*}{$50\%$}
 & OminiControl2 & \underline{8.47s} & \underline{35.8s} & \underline{113s} & \underline{286s} \\
 & FLUX-Fill & 21.5s & 113s & 383s & 1065s \\
 & ACE++ & 25.0s & 425s & --- & --- \\
 & EasyControl & 29.0s & 69.3s & 224s & 597s \\
 & \textbf{Ours} & \textbf{6.74s} & \textbf{19.2s} & \textbf{55.7s} & \textbf{173s} \\
\midrule
\multirow{5}{*}{$75\%$}
 & OminiControl2 & \underline{11.0s} & \underline{53.8s} & \underline{164s} & \underline{408s} \\
 & FLUX-Fill &  21.2s & 112s & 383s & 1052s \\
 & ACE++ & 25.0s & 426s & --- & --- \\
 & EasyControl & 29.2s & 69.4s & 221s & 587s \\
 & \textbf{Ours} & \textbf{8.32s} & \textbf{23.0s} & \textbf{84.1s} & \textbf{227s} \\ 
\bottomrule
\end{tabular}
\vspace{-0.1in}
\caption{Speed comparison of different methods across varying edit ratios and resolutions. Results marked by ``---" were unable to run on 96GB of GPU memory with expandable segments.}
\vspace{-0.15in}
\label{tab:speed-compare}
\end{table}



\subsection{Comparison with State-of-the-Art Methods}
We compare \ourwork with baselines from two different categories of editing, (1) instructional editing and (2) inpainting based editing. Instructional editing methods includes GPT Image 1~\cite{achiam2023gpt}, Gemini 2.5 Flash Image~\cite{team2023gemini}, FLUX.1 Kontext~\cite{blackforestlabs2024flux1dev} and SDEdit~\cite{meng2022sdedit}. For inpainting based editing we compare our method with FLUX-Fill~\cite{flux2024}, OminiControl2~\cite{tan2025ominicontrol2efficientconditioningdiffusion}, ACE++~\cite{mao2025ace++}, EasyControl~\cite{zhang2025easycontrol}.\footnote{\yy{See supplementary materials for evaluation metrics, benchmarks, and implementation details.}}

\vspace{2mm}
\noindent\textbf{Qualitative Results and Comparisons.}  
Figure~\ref{fig:teaser} showcases \ourwork's performance on diverse editing tasks at ultra-high resolutions (2K and 4K), demonstrating high visual fidelity and significant speed advantages over prior methods. Figure~\ref{fig:main-qualitative-instructional} presents comparisons on instruction-based edits using text-only or text--image prompts. In the first row, guided by both text and image, GPT Image 1 produces an object at the wrong scale, while FLUX.1 Kontext and SDEdit fail to generate the correct object, with SDEdit also introducing background distortions. In the second row for text-only edits, GPT Image 1 and FLUX Kontext fail to preserve the couch lighting, and Gemini and SDEdit significantly alter the global scene. 

Figure~\ref{fig:main-qualitative-inpainting} shows inpainting-based editing. In the first row, EasyControl fails to place the control image correctly, and ACE++ adds a window with incorrect color and pattern. We did not compare with OminiControl2 as their checkpoints are unavailable for subject-driven inpainting. In the second row, FLUX.1 Fill generates two glasses instead of one, EasyControl produces off-colored wine with shadow artifacts, and OminiControl2 and ACE++ place the lemon incorrectly inside the glass rather than beside it. 

\noindent\textbf{Quantitative Comparison.}  
Table~\ref{tab:Instruction-edit_full} reports quantitative results for instruction-based editing, where \ourwork achieves competitive performance across all benchmarks. Table~\ref{tab:inpainting_comparison_full} summarizes mask-guided and inpainting-based editing, showing that our approach consistently matches or outperforms baselines while offering substantial efficiency gains. Table~\ref{tab:speed-compare} presents average inference times for inpainting at different resolutions and edit ratios. \ourwork is the fastest across all configurations, with OminiControl2 consistently second. Notably, other methods often produce noisy or malformed outputs at high resolutions, making them inadequate for realistic edits. We provide full high-resolution stress tests in the supplementals, demonstrating that \ourwork is the only model capable of high-quality editing at these scales.

\section{High-Resolution Comparison}
\label{sec:high-resolution}
\yy{We compare our high-resolution generation results with those of other models, which fail to produce meaningful outputs. As shown in Figure~\ref{compare-res}, our method successfully generates high-quality 4K images, whereas the baselines either produce no content within the mask or fail entirely due to out-of-memory errors.}

\vspace{2mm}
\noindent\textbf{Efficiency Comparison.}
Table~\ref{tab:speed-compare} compares inference speed across methods. At $1\mathrm{K}$ resolution, our approach is up to $6.0{\times}$ faster than competing models, with even larger margins at higher resolutions. When varying the proportion of the edited region, our method remains highly efficient because computation is restricted to the modified areas, whereas competing approaches exhibit nearly constant runtime regardless of edit extent.

\begin{table}[t]
\centering
\small
\renewcommand{\arraystretch}{1.1}
\setlength{\tabcolsep}{5pt}
\begin{tabular}{l c c c c }
\toprule
\textbf{Method} & \textbf{Speed}$\downarrow$ & \textbf{PSNR}$\uparrow$ & \textbf{CLIP-T}$\uparrow$ & \textbf{CLIP-I}$\uparrow$ \\
\midrule
Ours & \textbf{2.34} & \underline{19.01} & \textbf{0.339} & \textbf{0.931} \\
-LWA & \textit{4.98} ($\times2.13$) & 18.84 & 0.336 & 0.922  \\
-LWA-FC & 9.32 ($\times3.98$)& 18.96 & \underline{0.338} & \underline{0.923}  \\
-LWA-FC-TI & 29.12($\times12.4$) & \textbf{19.03} & \textbf{0.339} & \underline{0.923} \\
\bottomrule
\end{tabular}
 \vspace{-0.1in}
\caption{Ablation of speed (without low-res  denoising part) difference on design components of our method on $1K\times1K$ resolution with $50\%$ edited region. LWA denotes local window attention with Flash Sparse Attention Kernel. The number in the () means how much slower comparing to the full pipeline.}
\vspace{-0.15in}
\label{tab:ablation-acceleration}
\end{table}


\section{Ablation Study}
\label{sec:ablation_study}

We conduct a series of ablation experiments to evaluate the contribution of key components in our pipeline, the need for the bounding box refinement, and the effect of step reduction during inference \yy{(see supplementary materials)}.

\vspace{2mm}
\noindent\textbf{Impact of Acceleration Components.}
Table~\ref{tab:ablation-acceleration} reports the effect of removing different acceleration modules. \textit{LWA} denotes the local window attention implemented with a Flash Sparse Attention kernel; \textit{TI} refers to the joint integration of conditional and noisy image tokens (as illustrated in the upper portion of Figure~\ref{fig:joint-denoise}); and \textit{FC} represents feature caching, which avoids redundant projections for unchanged regions. We evaluate four configurations: the full model, the model without LWA, without both LWA and FC, and without LWA, TI, and FC. Experiments are conducted at $1\mathrm{K}$ resolution with $50\%$ of the image edited. The results show that the complete system delivers the best trade-off between speed and fidelity, while progressive removal of components leads to notable degradation in both aspects.
\vspace{2mm}

\begin{figure}[h]
    \centering
    \includegraphics[width=0.95\linewidth]{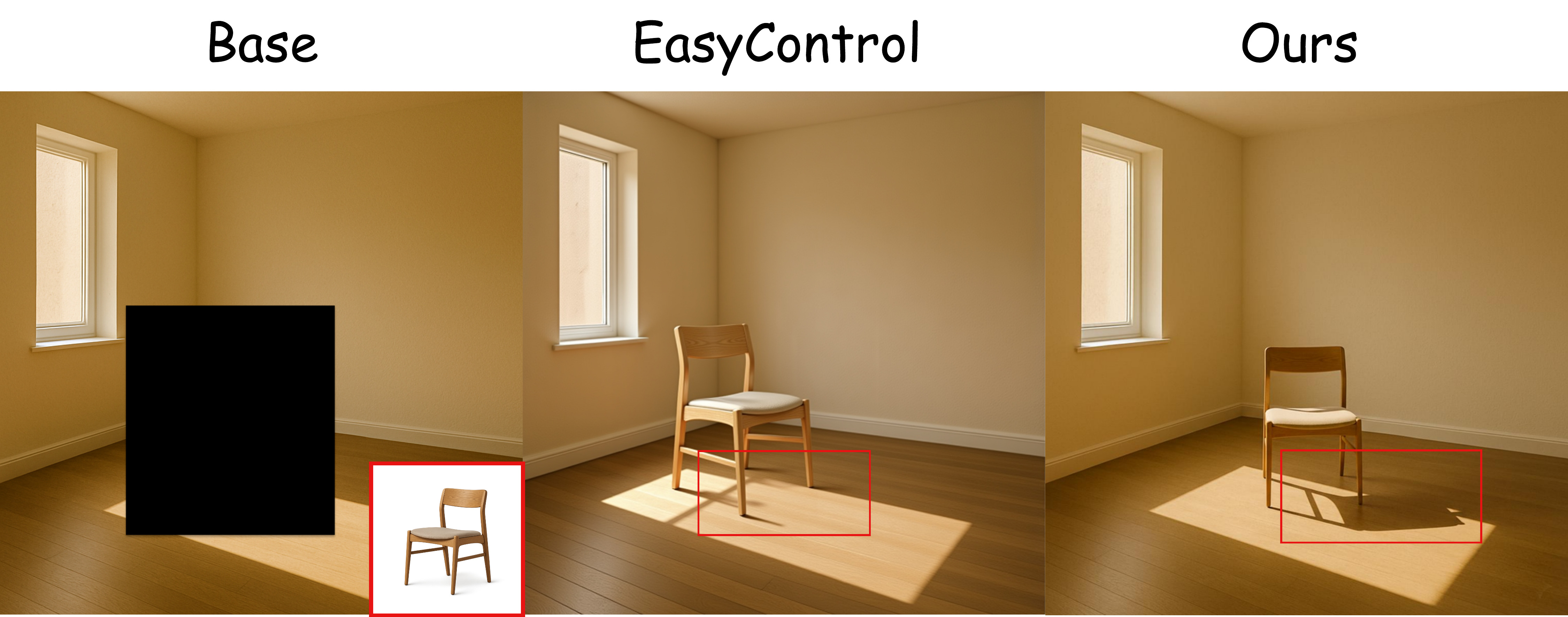}\\
    \vspace{-0.1in}
    \caption{The inappropriate bounding box will lead to issues in correct shadowing or other artifacts, therefore we need to refine the bounding box.}
    \vspace{-0.15in}
    \label{fig:chairm}
\end{figure}

\noindent\textbf{The Need for Bounding Box Refinement.}
Figure~\ref{fig:chairm} demonstrates the importance of bounding box refinement. In models like EasyControl, the direct usage of the given bounding box causes incorrect shadowing while ours solve this problem.

\section{Conclusion}
\label{sec:conclusion}
We presented \ourwork, a region-aware hierarchical diffusion framework for efficient high-resolution image editing, without requiring ultra-high-resolution training data. Our approach leverages low-resolution proxy guidance, local-window MMDiT, and inference acceleration to process only modified regions with sparse local-window attention, achieving substantial speedups over state-of-the-art methods while preserving visual fidelity. By decoupling computational cost from image resolution, \ourwork enables $4K$ editing that was previously impractical without dense attention or costly high-resolution data. These results demonstrate that combining multi-scale guidance, sparse attention, and targeted acceleration provides a practical and scalable solution for professional workflows demanding ultra-high-resolution outputs.

\newpage
{
    \small
    \bibliographystyle{ieeenat_fullname}
    \bibliography{cvpr_2026}
}
\newpage
\clearpage
\setcounter{page}{1}
\maketitlesupplementary
\begin{figure*}[t]
    \centering
    \begin{minipage}{\linewidth}
        \centering
        \includegraphics[width=\linewidth]{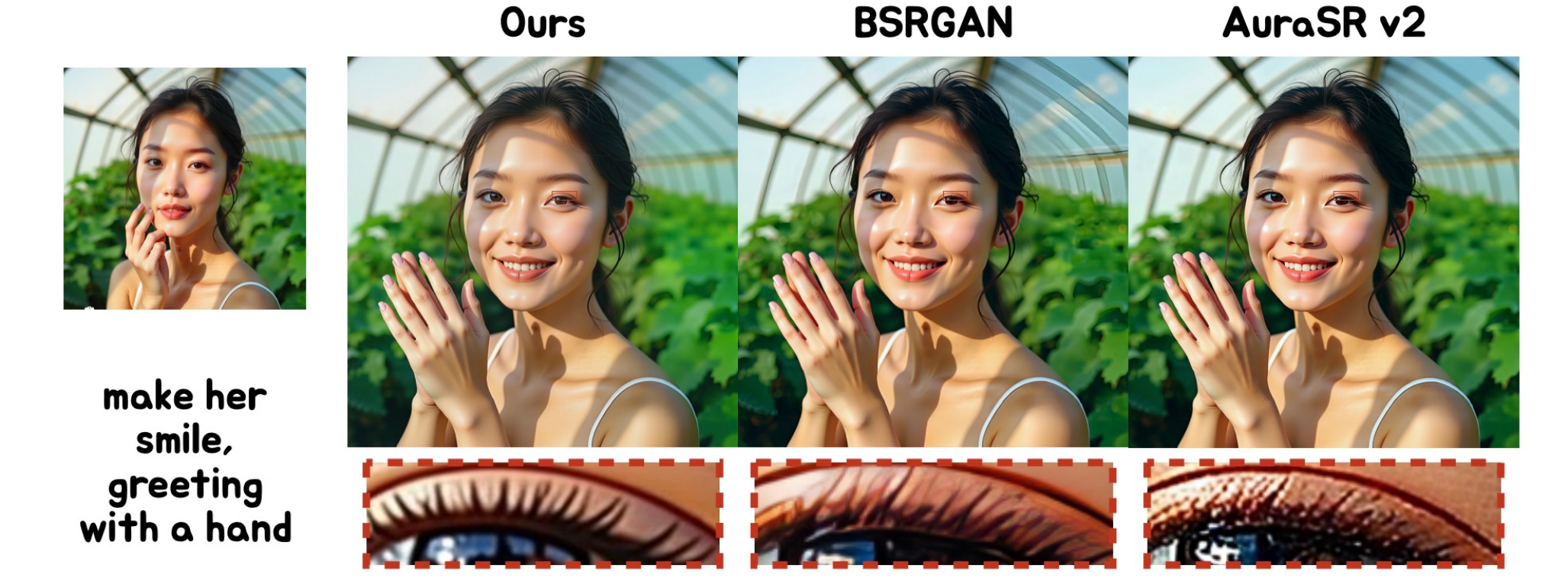}
        \vspace{-0.15in}
        \caption{Comparison with low-resolution editing followed by upscaling.}
        \label{fig:upscaler-compare}
    \end{minipage}
\end{figure*}
\yy{\section{Comparison with Upscaler Baselines.}}
\yy{Here we present a qualitative (Figure~\ref{fig:upscaler-compare}) example where the lady's facial expression and posture totally changed. Both upscalers show artifacts (eyelash discontinuity and blurriness), while ours does not. We will add complete quantitative comparisons in the final version. }

\section{Example on Full Editing}
\begin{figure}[h]
    \centering
    \includegraphics[width=0.95\linewidth]{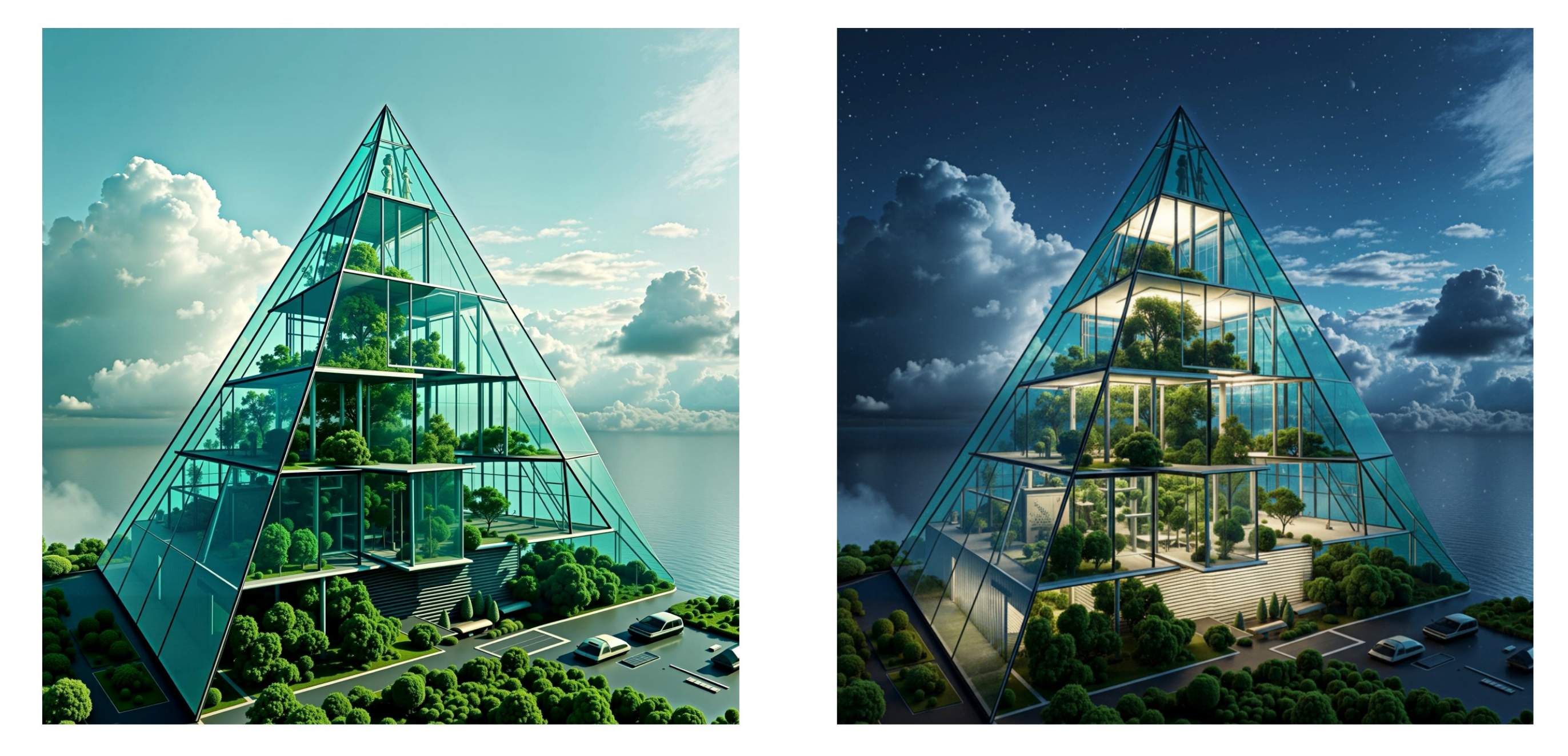}\\
    \vspace{-0.1in}
    \caption{Our model also supports full image editing by removing the mask although we focus on local editing.}
    \vspace{-0.15in}
    \label{fig:full-image-edit}
\end{figure}

\yy{Figure~\ref{fig:full-image-edit} further illustrates our ability to edit entire images, even though our focus is on local editing.}
\yy{\section{More Ablation Study}}
\begin{figure}[h]
    \centering
    \includegraphics[width=0.95\linewidth]{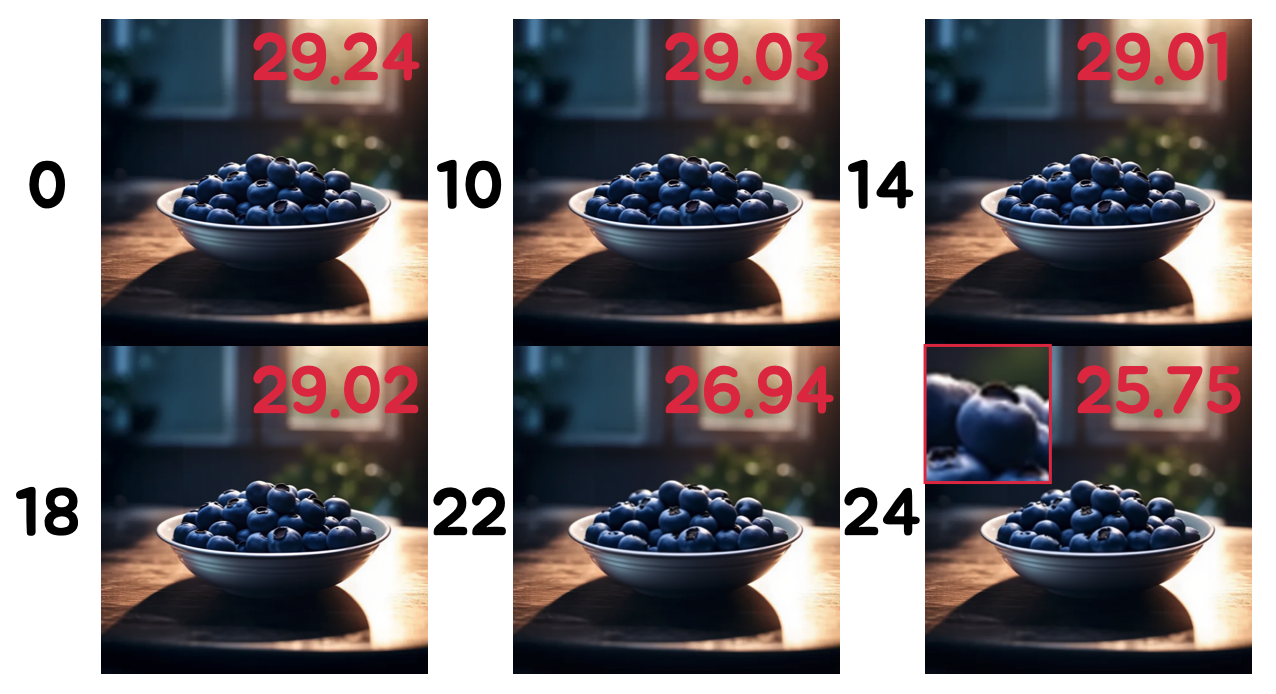}\\
    \vspace{-0.1in}
    \caption{Ablation on the number of timesteps skipped. Results shows 18 steps yields an optimal solution balancing the generation quality and speed. Numbers in red are PSNR values.}
    \label{fig:Ablation_timesteps}
    \vspace{-0.1in}
\end{figure}
\vspace{2mm}
\yy{\noindent\textbf{Ablation on Denoising Step Reduction.}
We further examine how aggressively denoising steps may be skipped while preserving output quality. Figure~\ref{fig:Ablation_timesteps} presents results obtained by skipping $\{0, 10, 14, 18, 22, 24\}$ out of 28 total steps. Skipping 18 steps offers the best balance between visual quality and efficiency.}

\begin{figure}[h]
    \centering
    \includegraphics[width=0.95\linewidth]{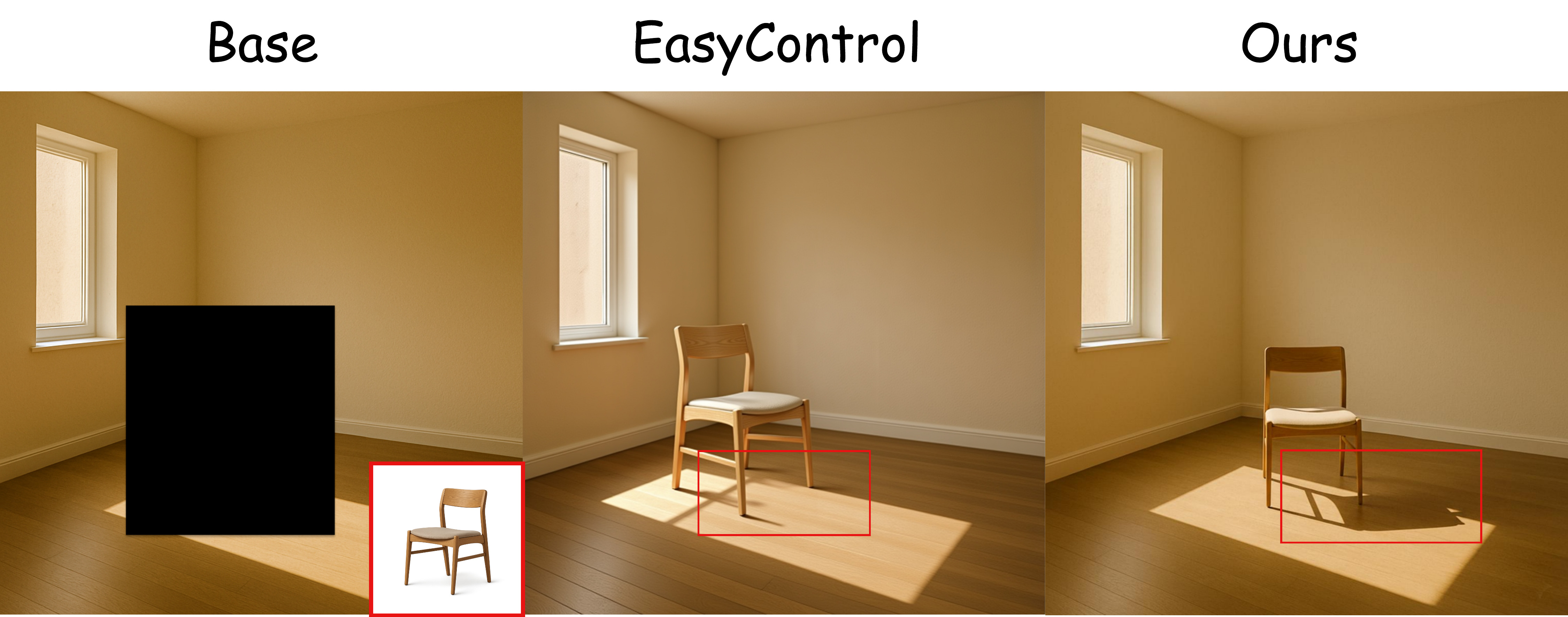}\\
    \vspace{-0.1in}
    \caption{The inappropriate bounding box will lead to issues in correct shadowing or other artifacts, therefore we need to refine the bounding box.}
    \vspace{-0.1in}
    \label{fig:chair}
\end{figure}

\vspace{2mm}

\section{Implementation Details}
\subsection{Experimental Setup.} We adopt FLUX.1-dev~\cite{flux2024} as our base model. The network is fine-tuned using LoRA with a rank of 16, trained for 10,000 steps with a batch size of 6 on six NVIDIA RTX A6000 Ada GPUs (48 GB each). All evaluations are performed on a NVIDIA RTX 6000 Ada Pro GPU (96 GB) server. For training, we employ the IPA300K dataset~\cite{zhang2025layercraft}, which provides paired source and target images along with textual prompts and bounding-box annotations on $1024\times1024$ resolution. For the off the shelf low-resolution model selection, we choose either FLUX-Kontext.dev or OminiControl2. \yy{For kernel adaptation, we permute the token sequence to a "window-first" way, so that the tokens within one local window will stick to each other in the sequence and we also adapt our code to Flash Sparse Attention~\cite{shi2025trainabledynamicmasksparse}, to skip the masked blocks for speeding up}
\subsection{Instructional Editing Benchmarks}
\label{sec:benchmarks}
We provide additional details regarding the quanitative comparisons in Table 1 in the main paper. We evaluate on I2EBench ~\cite{ma2024i2ebenchcomprehensivebenchmarkinstructionbased}, ImgEdit ~\cite{ye2025imgeditunifiedimageediting}, CompBench ~\cite{jia2025compbenchbenchmarkingcomplexinstructionguided}, and EmuEdit ~\cite{sheynin2023emueditpreciseimage}. For the composite benchmarks (those measuring multiple subtasks), we average the scores of the local editing tasks. The details for each benchmark are listed below.

\vspace{2mm}
\noindent\textbf{I2EBench.}
The I2EBench benchmark, which encompasses 16 diverse image editing tasks spanning both low-level restoration and high-level semantic modifications. We evaluate on the the first category includes 9 low-level tasks: Deblurring, HazeRemoval, Lowlight, NoiseRemoval, RainRemoval, ShadowRemoval, SnowRemoval, WatermarkRemoval, and RegionAccuracy. We evaluate using the Structural Similarity Index (SSIM), comparing edited images against ground truth references. 

\vspace{2mm}
\noindent\textbf{CompBench.}
We evaluate on the local subset of the  CompBench benchmark, which includes add, remove, and replace tasks. We report the average scores across those three tasks. For each task, we measure text-image alignment through the CLIP Score and structural similarity in background regions using SSIM.

\vspace{2mm}
\noindent\textbf{EmuEdit.}
For the EmuEdit benchmark, we utilize the test split of the facebook/emu\_edit\_test\_set dataset from HuggingFace, and report two metrics. The $\text{CLIP}_{\text{dir}}$ assesses directional alignment of the edit. The DINO score measures feature-level preservation.

\vspace{2mm}
\noindent\textbf{ImgEdit.}
We evaluate our method on the ImgEdit benchmark using its basic suite, which consists of single-turn editing tasks. The primary evaluation relies on a GPT-4 judge. The judge model receives three inputs (the original image, the edited image, and the textual instruction) and produces three integer scores ranging from 1 to 5 for different aspects of the edit. These scores are averaged. We report these averages across a range of local editing tasks: add, adjust, compose, extract, remove, and replace.

\section{More Examples}
\label{sec:more-examples}
\begin{figure*}[htbp]
    \centering
    \includegraphics[width=\linewidth]{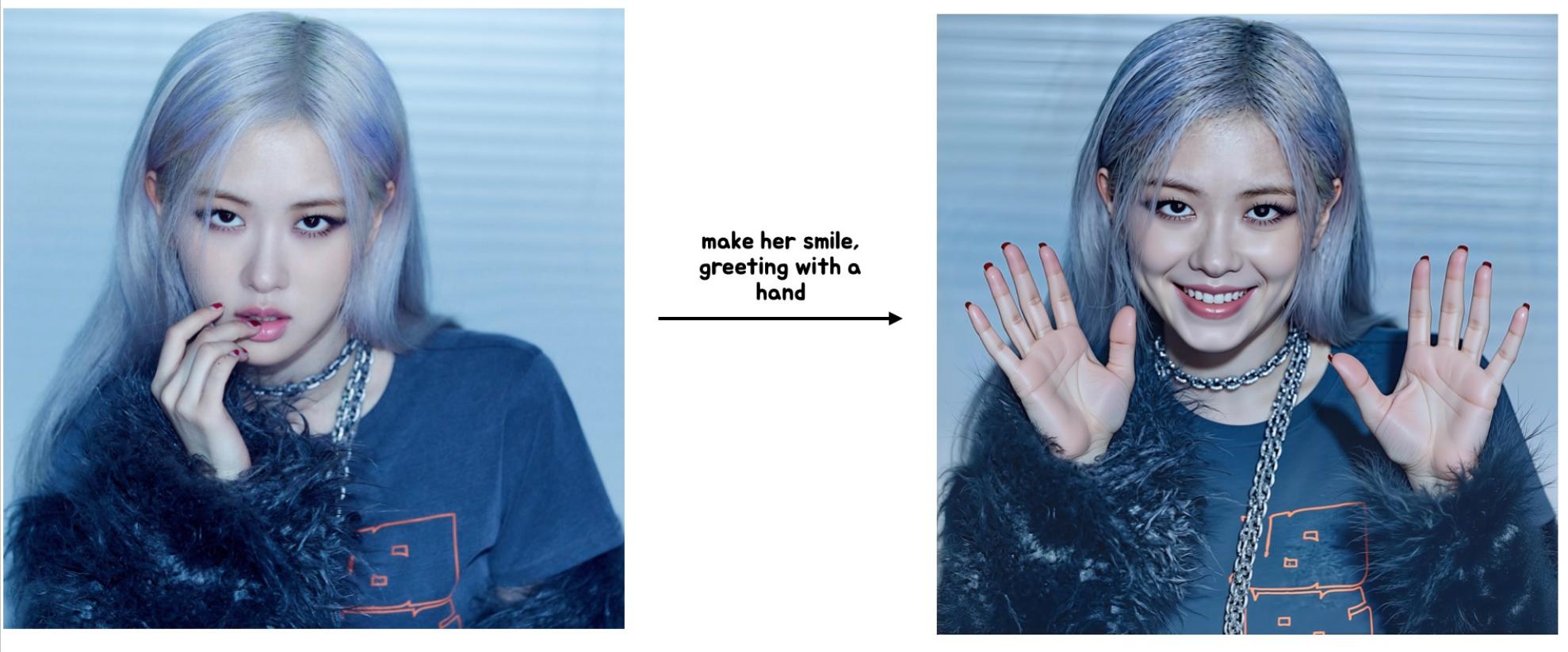}
    \includegraphics[width=\linewidth]{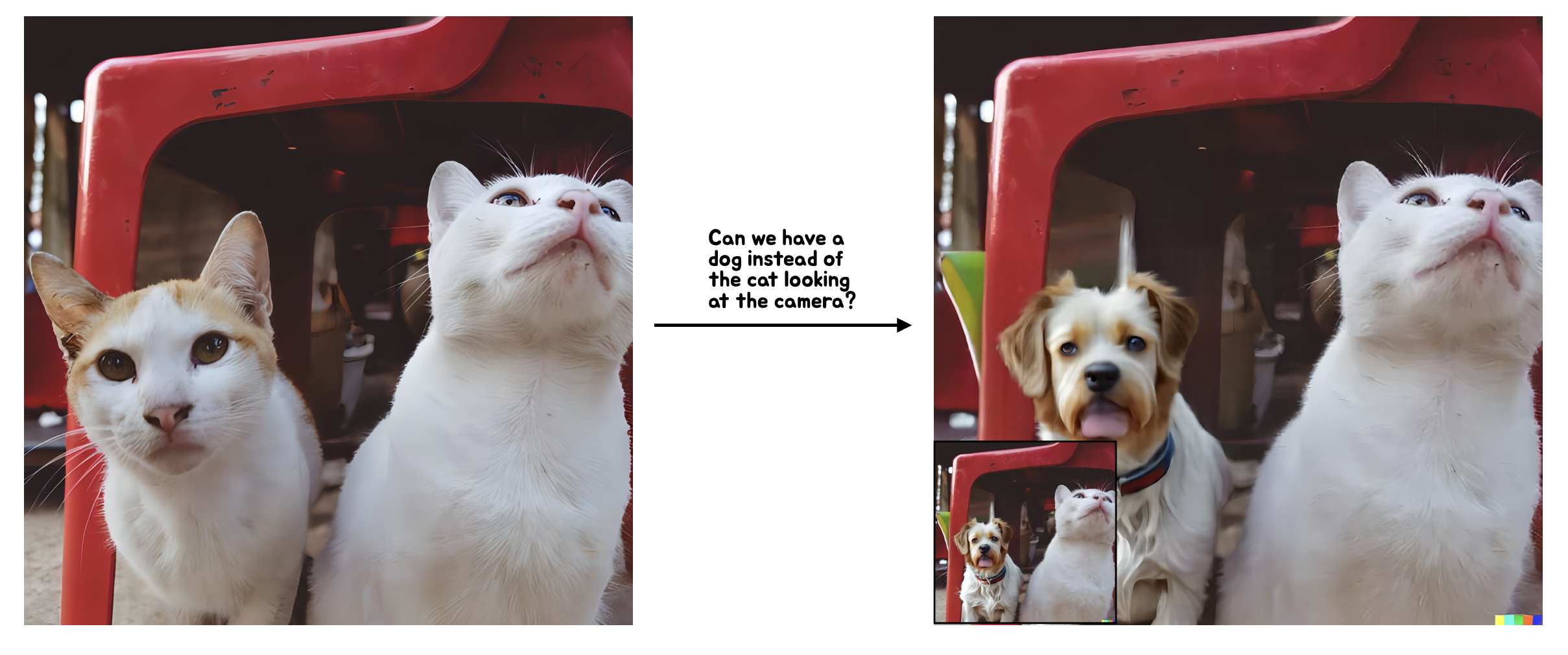}
    \includegraphics[width=\linewidth]{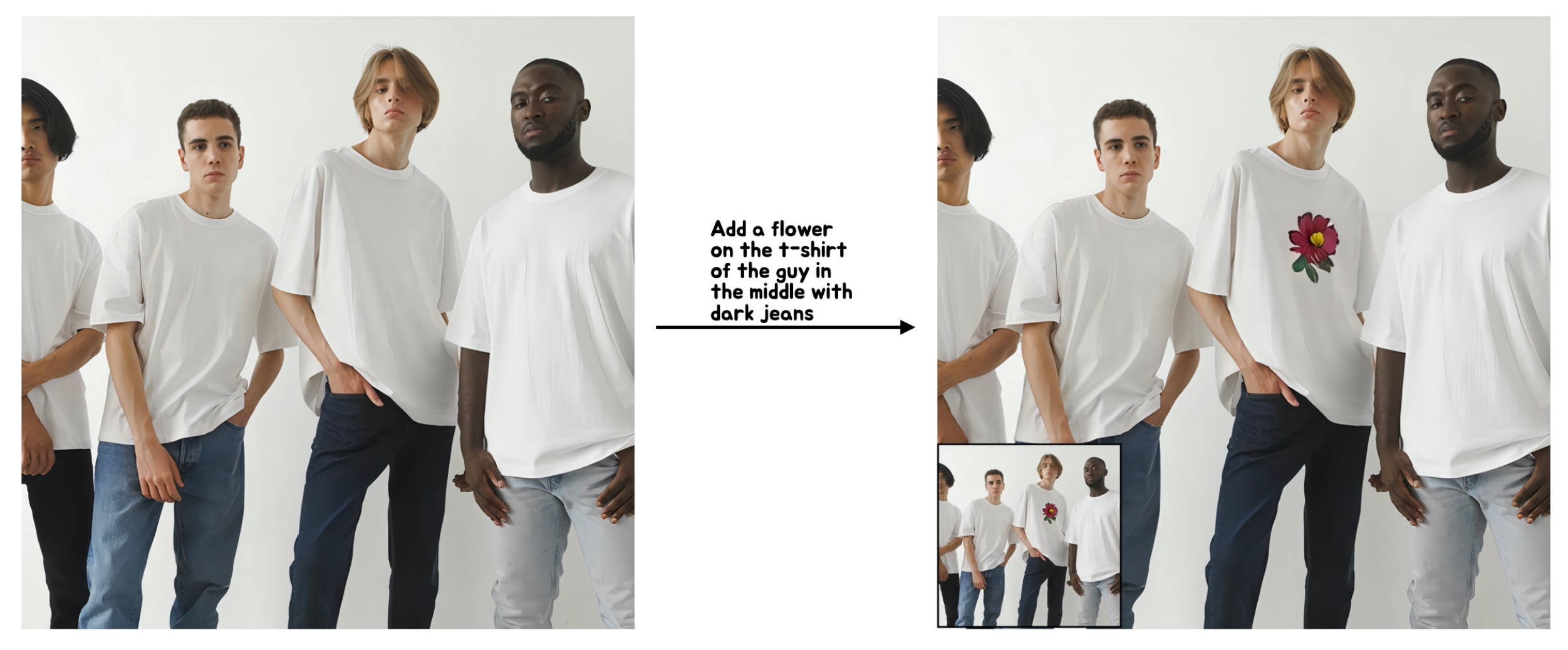}
    \caption{More examples on 4K in-the-wild data.}
    \label{fig:more-a}
\end{figure*}

\begin{figure*}[htbp]
    \centering
    \includegraphics[width=\linewidth]{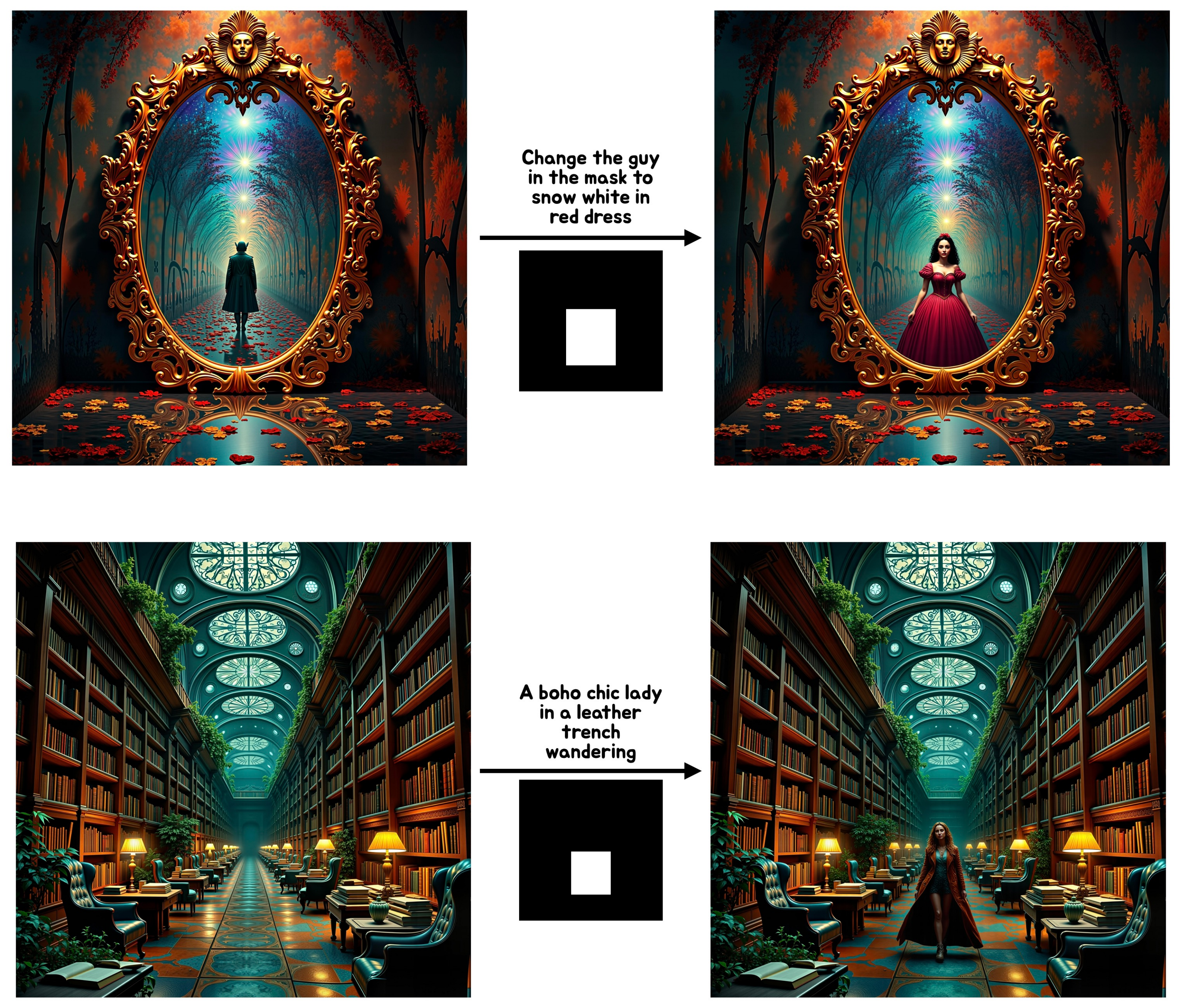}
    \caption{More examples on 4K synthetic data.}
    \label{fig:4k-more}
\end{figure*}

\begin{figure*}[htbp]
    \centering
    \includegraphics[width=0.95\linewidth]{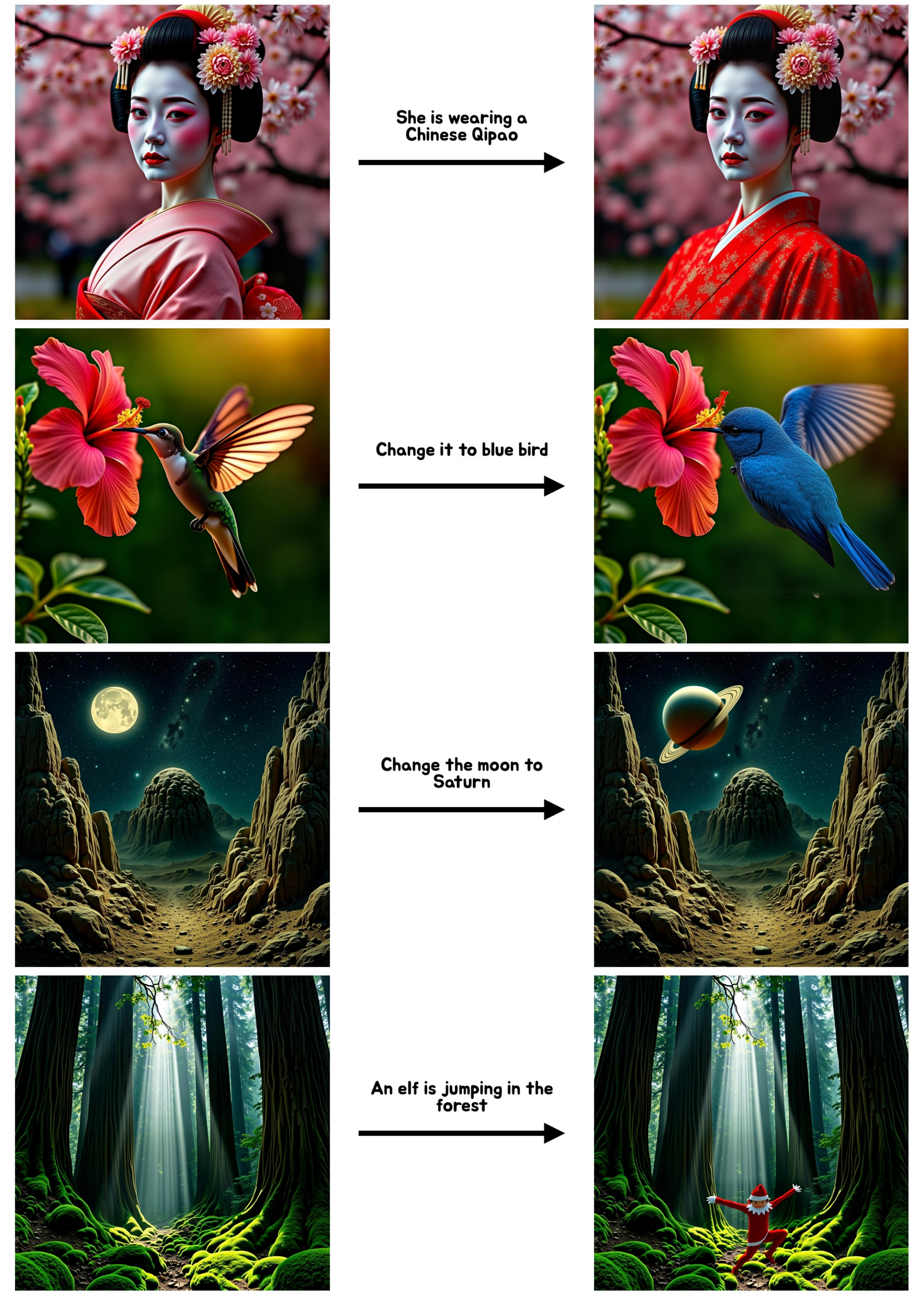}
    \caption{More examples on 2K synthetic data.}
    \label{fig:2k-more}
\end{figure*}

We present additional qualitative examples in Figures~\ref{fig:more-a}. The original images are upsampled using the BSRGAN super-resolution model~\cite{zhang2021designing}. For each example, the low-resolution edited result is displayed inside the black box located at the lower-left corner. We also include 4K (Figure~\ref{fig:4k-more}) and 2K (Figure~\ref{fig:2k-more}) editing results on synthetic data.

\if remove
\section{Pixel-Level Comparison}
\label{sec:pixel-comparison}

We perform a pixel-level comparison of the methods to demonstrate how they handle unedited (out-of-mask) regions. Specifically, we calculate the MSE for all out-of-mask pixels for a set of random samples for both text-guided inpainting and a set of random samples for image-guided inpainting. As the MSE is completely dependent on the specific sample set for each task, we only aim to compare the relative loss for each method to evaluate adherence to the base image. Table~\ref{tab:pixel-comparison} shows the result, where we are second best in the text-guided editing and best in the image-guided editing. One reason why we are second best in the text-guided editing is that we refine the mask to obtained a more appropriate place for the inpainting objects, thereby leading to more changes in the unmasked region.

\begin{table}[htbp]
    \centering
    \begin{tabular}{l|c c}
        \toprule
        \textbf{Method} & \textbf{Text-Guided} & \textbf{Image-Guided}\\
        \midrule
        FLUX-Fill & 20.79 & --- \\
        OminiControl2 & 18.52 & --- \\
        ACE++ & \textbf{10.74} & \underline{57.9} \\
        EasyControl & 45.92 & 72.5\\
        \textbf{Ours} & \underline{14.71} & \textbf{26.7} \\
        \bottomrule
    \end{tabular}
    \caption{Comparison of pixel MSE loss in the out-of-mask regions of a random sample of edited images. Results marked with ``---" do not support image-guided inpainting.}
    \label{tab:pixel-comparison}
\end{table}

\fi



\end{document}